\documentclass{article}

\usepackage{amsmath} 
\usepackage{algorithm}
\usepackage[noend]{algpseudocode}
\usepackage{graphicx}

\algnewcommand\algorithmicinput{\textbf{INPUT:}}
\algnewcommand\INPUT{\item[\algorithmicinput]}

\makeatletter
\def\BState{\State\hskip-\ALG@thistlm}
\makeatother

\begin{document}
\title{A Probabilistic Approach to Satisfiability of Propositional Logic Formulae}
\author{Reazul Hasan Russel 
	\\ rrussel@cs.unh.edu\\
	Department of Computer Science \\
	University of New Hampshire}
\date{}
\maketitle{} 


\section{Abstract}
We propose a version of {\em WalkSAT} algorithm, named as {\em BetaWalkSAT}. This method uses probabilistic reasoning for biasing the starting state of the local search algorithm. {\em Beta} distribution is used to model the belief over boolean values of the literals. Our results suggest that, the proposed {\em BetaWalkSAT} algorithm can outperform other uninformed local search approaches for complex boolean satisfiability problems.

\section{Introduction}
A statement that has either a true or false value is called a proposition. Propositional logic studies the idea of building complicated propositions using simple propositions or statements connected by logical connectives. The true or false value of a compound proposition depends on the values of underlying simple propositions and their structure of connectives. 

Two popular algorithms for determining the satisfiability of propositional logic are: {\em Davis–Logemann–Loveland (DLL)} algorithm and {\em WalkSAT} algorithm. {\em DLL} is a backtrack based complete algorithm and {\em WalkSAT} is an incomplete local search algorithm for determining the boolean satisfiability of the propositional logic formulae. These algorithms are unable to apply any prior belief about the assignments of the propositions. The {\em WalkSAT} algorithm always starts and continues with a uniform distribution and is unable to update it's belief about the variable assignment from an unsatisfiable situation, rather it repeatedly samples from the uniform distribution on each trial. The current algorithms don't employ the belief update or probabilistic reasoning about the binary assignment of the variables. This paper addresses the problem to make the {\em WalkSAT} algorithm smarter by updating the belief of variable assignments with each unsatisfied attempt, so that it can obtain a positive bias towards convergence and can find a satisfiable or unsatisfiable situation more efficiently.

The paper is organized as follows: section 2 discusses some of the relevant works. Section 3 presents the approach and the proposed algorithm. Section 4 presents and discusses the results obtained. Section 5 concludes the paper with some probable future works.

\section{Related works}
Search strategy can be divided broadly into two groups: complete and heuristic \cite{Bengoetxea2003}. A complete strategy considers all possible options to find a solution whereas a heuristic based strategy prunes out the less promising options to quickly reach the goal. Heuristic strategies can be deterministic or non-deterministic. A deterministic strategy ({\em e.g. different hill climbing algorithms}) yields the same solution under the same condition and can get stuck into the local maxima or minima. A non-deterministic strategy is able to produce different solutions under the same condition and can escape from local minima or maxima by the strength of it's stochasticity. Evolutionary algorithms, Genetic algorithms, Simulated annealing etc. utilizes the non-deterministic strategy. But these non-deterministic strategy approaches have some shortcomings like: they are not guaranteed to always find an optimal solution, can be very slow and complex, depends on many parameters that require manual tuning and finding the optimal parameters itself yields an optimization problem to solve \cite{Tabassum2014, Boese1994}.

Estimation distribution algorithms (EDA) biases the movement of the search towards the more promising candidate options by maintaining an explicit probabilistic model. EDA is an evolutionary algorithm also known as probabilistic model-building genetic algorithm that treats the optimization problem as a sequence of incremental update of probabilistic model starting from a prior belief model ({\em e.g. uniform}) \cite{Pelikan2012}. The incremental bias towards the goal makes EDA a good candidate to solve complex optimization problems. There are several papers available on EDA containing its application in bioinformatics, genomics, DNA microarray classification, proteomics and so on \cite{Armananzas2008}. A mixed random walk strategy is proposed in \cite{Selman1995} for escaping from the local minima for satisfiability testing of CNF formulas. \cite{Zhang2001} presents different conflict driven learning techniques used in various SAT solvers and proposes several new learning strategies. To the best of our knowledge, we did not find a paper that applies the idea of Estimation distribution algorithms for boolean satisfiability of propositional logic, maintaining the explicit probabilistic model as a {\em beta distribution} of the binary literals.

\section{Approach}
In probability theory, {\em Bernoulli distribution} is the probability distribution of a random variable that has a value of {\em true} with probability {\em p} and {\em false} with probability {\em 1-p}. The {\em WalkSAT} algorithm uniformly samples a true value for a proposition with probability {\em 0.5} and a false value with probability {\em 0.5}.  When a specific assignment to variables lead to an unsatisfiable situation, there is a reason to change the belief that the value of a currently assigned proposition is uniformly distributed. If a variable is assigned to true and the final outcome is unsatisfiable, we may change our belief a little that a false assignment to this proposition might have a better chance to lead to satisfiability.

\begin{algorithm}
	\caption{Beta-WalkSAT algorithm applying Estimation distribution algorithms strategy with positive bias for starting state using explicit probabilistic model with {\em Beta} distribution}\label{Beta-WalkSAT}
	\begin{algorithmic}[1]
		\Procedure{ProbabilisticWalkSAT}{}\\
		\textbf{Input: } clauses, a set of clauses in propositional logic\\
		\qquad \qquad p, a probability to choose random walk\\
		\qquad \qquad maxTries, number of maximum tries\\
		\qquad \qquad maxFlips, number of maximum flips\\
		\textbf{Output: } A solution if consistent, false otherwise\\
		\For{\texttt{1 to maxTries}}
		\State \texttt{assign all variables sampling from the {\em beta} distribution\\\qquad \qquad of the explicit probabilistic model of EAD}
		
		\For{\texttt{1 to maxFlips}}
		
		\State \texttt{randomly choose an unsatisfied clause c}
		\If{one or more of c's variables can be flipped while breaking \\ \qquad \qquad \quad nothing}
		\State randomly chose among those
		
		\Else
		\State with probability p\\
		\qquad \qquad \qquad \qquad randomly chose one of c's variables
		\State \textbf{else}\\
		\qquad \qquad \qquad \qquad randomly choose among those of c's variables that\\
		\qquad \qquad \qquad \qquad minimize breaks.
		\EndIf
		
		\State flip the variable
		\State if formula satisfied, terminate
		\EndFor{\textbf{end for}}
		\State \texttt{update explicit probabilistic model of {\em beta} distribution\\
			\qquad \quad of the Random variables (literals) based on EAD}
		\EndFor{\textbf{end for}}
		\EndProcedure
	\end{algorithmic}
\end{algorithm}

{\em Beta distribution} is a very convenient and powerful choice of priors for Bernoulli distribution. The beta distribution presents a continuous probability distribution of random variables on the interval [0,1]. Beta distribution is parametrized by two positive parameters $\alpha$ and $\beta$ that controls the shape of the distribution.

The algorithm presented here assumes the literals are independent of each other, they are random variables with binary choices and  models the belief of the assignment of each literal with a beta distribution. Initially it represents the prior belief about the assignment which is uniform. After each unsuccessful trial, the belief for each literal is updated with the beta distribution. Sampling for the next trial is then done from this distribution, biasing the chance towards faster convergence.

\subsection{Beta-WalkSat algorithm}
The {\em Beta-WalkSat} algorithm described in {\em algorithm 1} is almost same as of the normal {\em WalkSAT} algorithm except that it maintains a beta distribution for each literal and utilizes that distribution for biasing the starting state of each trial. Lines 2-5 describes the input to the algorithm and line 6 describes the returned output.

The algorithm starts at line 8 and iterates for {\em maxTries} number of times. At the start of each trial, an initial value is assigned to each literal sampling from the {\em beta} distribution of that literal in lines 9-10. The for loop at line 11 runs for {\em maxFlip} number of times, picks a random unsatisfied clause in line 12, tries to find a literal that doesn't break any other clause and flips that in lines 13-15. If no such literal is found, it flips a random literal with probability p in line 18, or with probability 1-p chooses a random literal that minimizes breaks. If all the clauses are satisfied, the result is returned in line 23. If not, it updates the beta distribution of each literal by increasing the value of alpha or beta based on the current assignment of the literal in lines 24-25 and proceeds for the next trial.

\section{Result}
The result comparison is done among four different versions of WalkSAT algorithms. The {\em BetaWalkSAT} algorithm is the one that's described above. The {\em 5BestWalkSat} and {\em AllWalkSAT} algorithms are introduced for comparison purpose to see how it behaves compared to some random variations of the general WalkSAT algorithm. The {\em 5BestWalkSat} algorithm stores the variable assignments of 5 best previous trials and samples the initial assignment of next trial from this information. The {\em AllWalkSAT} version works same as {\em 5BestWalkSat}, but it keeps track of all previous literal assignments instead of 5.

\begin{figure}[ht!]
	\centering
	\includegraphics[width=\textwidth]{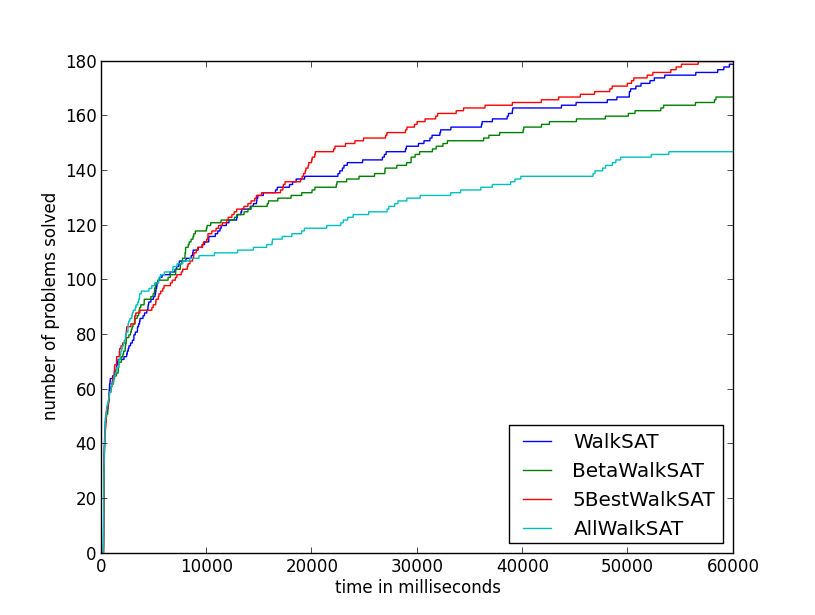}
	\caption{\small \sl number of solved problems vs. time \label{fig:solved_vs_time}}
\end{figure}

Figure \ref{fig:solved_vs_time} presents the number of problems solved in X-axis and the time in Y-axis. The presented {\em BetaWalkSAT} algorithm has to sample from {\em Beta} distribution of each variable, which involves calculating the inverse of Cumulative Distribution Function (CDF). This is an expensive operation making the {\em BetaWalkSAT} algorithm a little slower. The {\em WalkSat} and {\em 5BestWalkSat} algorithms doesn't involve any complex calculation and thus works faster in simple solvable problems. The graph could be presented as number of problems solved vs. number of flips required, but as the number of tries are very few for the simple satisfiable problems and number of flips highly varies randomly, that seems not a reasonable way to express the result.

	\begin{figure}[ht!]
		\centering
		\includegraphics[width=\textwidth]{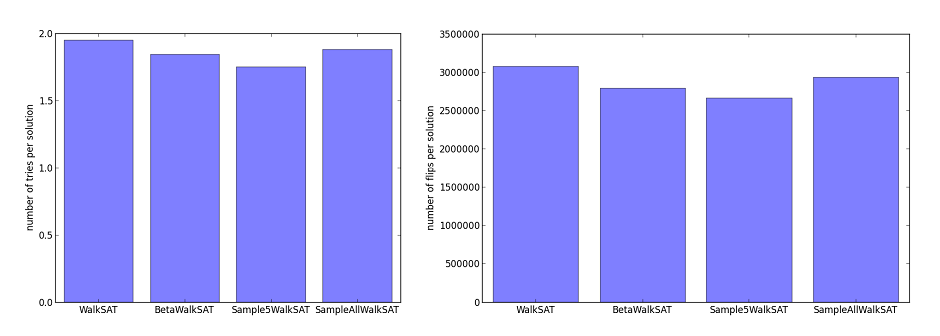}
		\caption{\small \sl average number of tries and flips per solution \label{fig:tries_flips}}
	\end{figure}
	
Figure \ref{fig:tries_flips} shows the average number of tries and flips per solution. The available CNF problem generators generate problems that are either easily satisfiable or are not satisfiable at all. They generally do not generate problems that are satisfiable, but are complex and difficult to solve. Though thousands of CNF problems are generated for experimental purposes, the satisfiable problems among them are very easily solvable yielding very few number of tries per solution. As the intelligence of the {\em BetaWalkSAT} algorithm comes into play when sampling for the starting state is performed for a new trial, the generated samples using the CNF generators doesn't help much in evaluating the {\em BetaWalkSAT} algorithm.

	\begin{figure}[ht!]
		\centering
		\includegraphics[width=\textwidth]{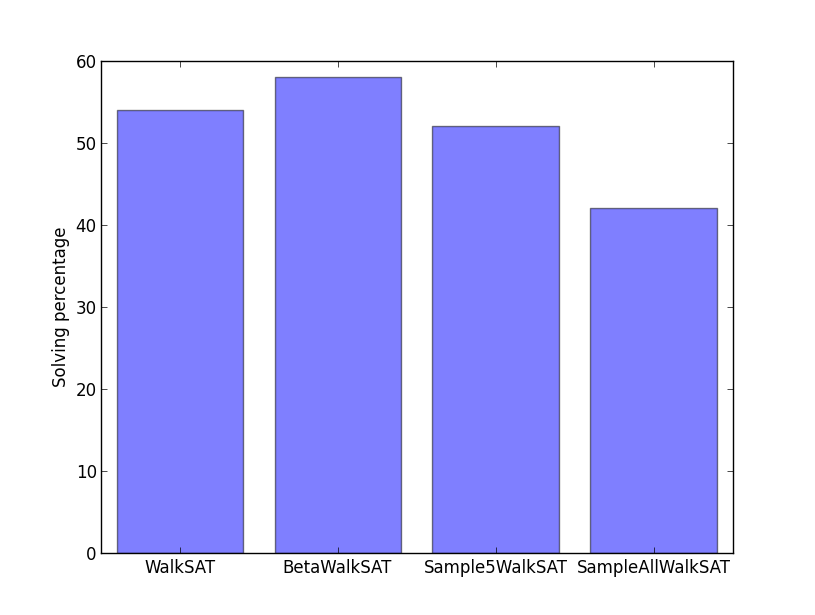}
		\caption{\small \sl solving percentage of each algorithm for a hard satisfiable problem  \label{fig:solve_percentage}}
	\end{figure}
	
Figure \ref{fig:solve_percentage} presents the percentage of satisfiable runs of each algorithm for a complex satisfiable problem. {\em BetaWalkSAT} performs better in this case bolstering the idea of giving positive bias to the initial assignment of literals at the start of each trial. As the problem considered here is complex, it requires many trials to find a solution and {\em BetaWalkSAT} algorithm is able to apply its reasoning for the positive bias of the starting state.

\section{Conclusion}
The starting state is critically important for any local search algorithm. Finding a global minima or maxima can extensively be affected depending on where the search started. {\em WalkSAT}, the existing local search algorithm for determining the satisfiability of propositional logic formulae is not able to bias the starting state towards more promising states. The proposed {\em BetaWalkSAT} algorithm improves the starting state over consecutive trials biasing towards convergence. The results presented in the paper bolsters the claim showing that, biasing starting state by maintaining {\em Beta} distribution for each literal does help in determining satisfiability. It outperforms for harder problems requiring many trials, though the difference is not significant enough. Applying {\em BetaWalkSAT} in more complex solvable problems would help to better understand its performance. In future, we will work more to formulate complex solvable problems to better quantify the performance improvement of our algorithm. The idea of biasing starting state is general and is supposed to work for other local search algorithms as well. In future, we will work to see how it behaves for other local search algorithms.

\section{Acknowledgment}
Special thanks to professor Wheeler Ruml for the awesome AI course and for all the great helps and suggestions regarding this project.

\bibliographystyle{plain}
\bibliography{reazullib}

\end{document}